\documentclass[sn-mathphys,Numbered]{sn-jnl}

\usepackage{graphicx}%
\usepackage{multirow}%
\usepackage{amsmath,amssymb,amsfonts}%
\usepackage{amsthm}%
\usepackage{mathrsfs}%
\usepackage[title]{appendix}%
\usepackage{xcolor}%
\usepackage{textcomp}%
\usepackage{manyfoot}%
\usepackage{booktabs}%
\usepackage{algorithm}%
\usepackage{algorithmicx}%
\usepackage{algpseudocode}%
\usepackage{listings}%

\theoremstyle{thmstyleone}%
\theoremstyle{thmstyletwo}%

\theoremstyle{thmstylethree}%

\begin{document}

\title[Article Title]{OPSD: an Offensive Persian Social media Dataset and its baseline
evaluations}


\author*[1]{\fnm{Mehran} \sur{Safayani}}\email{safayani@iut.ac.ir}

\author[2]{\fnm{Amir} \sur{Sartipi}}\email{amirsartipi.msc@eng.ui.ac.ir}

\author[1]{\fnm{Amir Hossein} \sur{Ahmadi}}\email{amir\_ahmadi@ec.iut.ac.ir}

\author[1]{\fnm{Parniyan} \sur{Jalali}}\email{p.jalali@ec.iut.ac.ir}

\author[1]{\fnm{Amir Hossein} \sur{Mansouri}}\email{a.mansouri@ec.iut.ac.ir}

\author[3]{\fnm{Mohammadamir} \sur{Bisheh-Niasar}}\email{ma\_bishehniasar@ie.sharif.edu}

\author[4]{\fnm{Zahra} \sur{Pourbahman}}\email{pourbahman@aut.ac.ir}

\affil[1]{\orgdiv{Department of Electrical and Computer Engineering}, \orgname{Isfahan University of Technology}, \orgaddress{\city{Isfahan} \postcode{84156-83111}, \country{Iran}}}

\affil[2]{\orgdiv{Faculty of Computer Engineering}, \orgname{University of Isfahan}, \state{Isfahan}, \country{Iran}}

\affil[3]{\orgdiv{Department of Industrial Engineering}, \orgname{Sharif University of Technology}, \orgaddress{\city{Tehran}, \country{Iran}}}

\affil[4]{
\orgdiv{Computer Engineering Department}, 
\orgname{Amirkabir University of Technology (Tehran Polytechnic)}, \orgaddress{No. 350, Hafez Ave, Tehran 1591634311, Iran}}


\abstract{
\textbf{Warning} This paper includes a sample of comments extracted from the dataset, some of which may contain offensive or obscene content.

The proliferation of hate speech and offensive comments on social media has become increasingly prevalent due to user activities. Such comments can have detrimental effects on individuals' psychological well-being and social behavior. While numerous datasets in the English language exist in this domain, few equivalent resources are available for Persian language. To address this gap, this paper introduces two offensive datasets. The first dataset comprises annotations provided by domain experts, while the second consists of a large collection of unlabeled data obtained through web crawling for unsupervised learning purposes. To ensure the quality of the former dataset, a meticulous three-stage labeling process was conducted, and kappa measures were computed to assess inter-annotator agreement. Furthermore, experiments were performed on the dataset using state-of-the-art language models, both with and without employing masked language modeling techniques, as well as machine learning algorithms, in order to establish the baselines for the dataset using contemporary cutting-edge approaches. The obtained F1-scores for the three-class and two-class versions of the dataset were 76.9\% and 89.9\% for XLM-RoBERTa, respectively.}




\maketitle

\section{Introduction}\label{sec1}

Social media platforms have experienced a surge in popularity in recent years, providing users with a platform to express their opinions across various forms of media. Among these forms, textual data holds significant importance as a means of conveying meaning and ideas. Unfortunately, it is not uncommon for certain individuals to incorporate abusive language within their sentences, resulting in offensive comments. Detecting such instances manually would be a time-consuming and expensive endeavor. However, thanks to advancements in Natural Language Processing (NLP) applications, it has become possible to automatically identify and classify these comments as offensive or non-offensive. This task can be approached as a classification problem, requiring a suitable dataset consisting of sentences of this nature. Regrettably, to the best of our knowledge, there are only few datasets available for offensive comment detection. Leveraging these NLP applications can prove instrumental in filtering offensive content from social media platforms, particularly for children or individuals who prefer not to be exposed to such material.

In this study, we collected approximately 5,000 and 17,000 comments from two widely used social media platforms, namely Instagram and Twitter, respectively. To ensure accurate labeling of the data, a meticulous three-phase annotation process was employed. In addition, a collection of 170,000 unlabeled data points in Persian was gathered using offensive keywords and hashtags. The primary contributions of this study can be summarized as follows:

\begin{enumerate}
  \item 
Introduction of the offensive Persian Social Media Dataset (OPSD).
  \item 
    Investigation of the impact of masked language modeling on pre-trained language models using the unlabeled portion of the dataset.
  \item 
    Comprehensive experiments were conducted on the annotated dataset, involving transformer-based language models and machine learning approaches.
\end{enumerate}

The remaining sections of this article are organized as follows: Section \ref{sec:related-work} reviews previous studies relevant to datasets in the field. Section \ref{sec:data-collection} explains the data collection process in detail. Section \ref{sec:annotation} discusses the annotation methodology employed. Section \ref{sec:experiments} presents the extensive experiments conducted and the establishment of baselines for the dataset using transformer-based language models and machine learning approaches. Finally, Section \ref{sec:conclusion} presents the concluding remarks.

\section{Related Work}
\label{sec:related-work}

In this section, we first review prior work on offensive
language datasets in the English language and then
present the related literature on offensive datasets in the Persian language.

\subsection{English Offensive Datasets}

The OLID \citep{zampierietal2019} dataset is designed for the detection (Subtask-A), categorization (Subtask-B), and identification of offensive language (Subtask-C). Subtask-A focuses on binary classification with 'NOT' and 'OFF' labels. 'NOT' indicates posts without any offensive content, while 'OFF' refers to posts that include insults, threats, profanity, or swear words. The OLID dataset consists of 14,100 samples, with 13,240 samples for training and 860 samples for testing set.

The Semi-Supervised Offensive Language
Identification Dataset (SOLID) \cite{rosenthal-etal-2021-solid} was introduced to address the scarcity of offensive language instances in the OLID dataset and the time-consuming annotation process. SOLID utilized a semi-supervised approach, starting with the OLID dataset, to collect new offensive data. This novel dataset overcomes the limitations of existing datasets and provides a substantial collection of English tweets for offensive language identification. SOLID demonstrates improved offensive language detection capabilities and includes a comprehensive analysis of various types of offensive tweets. The hierarchical labeling schema used in SOLID and related works is based on the OLID dataset.

HSOL \cite{Davidson_Warmsley_Macy_Weber_2017} is a dataset created for the purpose of hate speech detection. The authors began by utilizing a hate speech lexicon, which consisted of words and phrases identified as hate speech by internet users and compiled by Hatebase.org. By using the Twitter API, they searched for tweets that contained terms from the lexicon. This process resulted in a sample of tweets collected from 33,458 Twitter users. The authors then extracted the timeline of each user, resulting in a corpus of 85.4 million tweets. From this corpus, a random sample of 25,000 tweets that contained lexicon terms was selected for manual coding by CrowdFlower (CF) workers. The workers were tasked with labeling each tweet into one of three categories: hate speech, offensive but not hate speech, or neither offensive nor hate speech.

\subsection{Persian Offensive Datasets}

The Persian Offensive Language Identification Dataset (POLID) is one of the pioneering datasets in this domain, developed by \cite{alavi2021offensive}. The dataset encompasses a diverse range of text data, including tweets, Instagram comments, and user reviews sourced from popular Iranian web applications such as Digikala and Snappfood. The annotation process for POLID involved a semi-automatic approach. Initially, a basic list of common swear words was used to label each text as either "OFF" (offensive) if it contained any of these words, or "NOT" (inoffensive) if it did not. Subsequently, the authors manually reviewed and corrected the labels assigned to the text data. POLID consists of 4,988 text entities, with 2,453 labeled as inoffensive and 2,535 as offensive.

\citet{mozafari:tel-03276023} employed both random and lexicon-based sampling methods to collect tweets from June to August 2020. In the random sampling approach, tweets were selected randomly and evaluated by two experts. The analysis revealed that only a maximum of 2\% of the selected tweets contained offensive content, resulting in an imbalanced sampling. On the other hand, the lexicon-based sampling involved the utilization of a word list to filter tweets, ensuring a more unbiased representation across different topics. The random sampling yielded 320,000 tweets, while the lexicon-based sampling resulted in 200,000 tweets. Approximately 6,000 of the collected tweets were manually annotated.

 The PerBOLD dataset \cite{perbold} focuses on the acquisition and annotation process for building a dataset aimed at automatic detection of offensive speech on Instagram. Two approaches were employed to collect textual data: a user-based approach targeting controversial users' pages and a news agencies-based approach focusing on comments published on news agency pages. Due to restrictions imposed by Instagram, data extraction was conducted partially from the platform itself and partially from the picuki website. The collected data underwent an annotation process, taking into consideration the subjective nature of offensiveness and the necessity for accurate tagging. A sample of approximately 30,000 comments was randomly selected and labeled by three annotators, with the involvement of a linguist expert to resolve any inconsistencies. The annotators classified comments as offensive, non-offensive, or advertisement, while offensive comments were further categorized into various types such as curse, insult, sexist, racist, and more. This work underscores the significance of precise tagging for effective machine learning classifiers. Overall, the paper highlights the challenges and methodologies involved in creating a dataset for offensive speech detection on Instagram.

Pars-OFF \cite{9936700} is a comprehensive annotated corpus comprising 10,563 data samples for offensive language detection in Persian. To ensure balance in the dataset, a combination of similarity-based and keyword-based data selection techniques were employed during the collection of tweets. Additionally, this paper presents an evaluation of traditional machine learning approaches and Transformer-based models on the Pars-OFF dataset as a baseline. The BERT+fastText model achieved the highest performance, attaining an impressive F1-Macro score of 89.57\%.

\section{Data collection}
\label{sec:data-collection}

In this section, we introduce OPSD and delve into how the dataset was collected and annotated. This section describes the data collection process for OPSD (Twitter), OPSD (Instagram), and also OPSD (Unlabeled), as demonstrated in Subsections \ref{OPSD(Twitter)}, \ref{OPSD(Instagram)}, and \ref{OPSD(Unlabeled)}, respectively.

\subsection{OPSD (Twitter)}
\label{OPSD(Twitter)}

To collect data from Persian Twitter, a list of keywords was compiled by inspecting trending hashtags and topics known to contain a higher percentage of offensive content. Over 40,000 records were extracted by searching for these keywords. An initial pre-processing step was conducted to remove records from company support accounts and eliminate duplicates. From each keyword, a subset containing more than 16,000 records was sampled. Finally, these samples were aggregated to form a dataset of 17,000 records.

\subsection{OPSD (Instagram)}
\label{OPSD(Instagram)}

The data collection process for OPSD (Instagram) involved crawling over 13,000 records from the Instagram platform. Specifically, a set of 100 comments from different posts was extracted from the accounts of an IT service company. Upon examination, it was discovered that a significant amount of the collected data was unusable. Therefore, an initial pre-processing step was carried out to remove records that contained less than one Persian word, after removing emojis, emoticons, hashtags, mentions, links, and all non-Persian text. Due to the diverse nature of the posts and the evenly distributed number of collected data from each post, a sample of 5,000 records was extracted from the crawled data.

\subsection{OPSD (Unlabeled)}
\label{OPSD(Unlabeled)}

The OPSD (Unlabeled) dataset was collected following a similar procedure as described in Subsection \ref{OPSD(Twitter)}. In this case, a collection of over 140,000 tweets were crawled from Twitter. An initial pre-processing step similar to previous datasets was applied. Finally, these records formed the OPSD (Unlabeled) dataset.

\section{Annotation}
\label{sec:annotation}

The OPSD dataset is annotated with three labels, which are defined as follows:

\begin{itemize}
\item \textbf{POS} (Positive): A comment is labeled as POS if it does not contain any destructive concepts, such as hate speech, offensive language, or abusive content.

\item \textbf{T-NEG} (Targeted Negative): Comments with the T-NEG label specifically target individuals, companies, or organizations with hateful and offensive content.

\item \textbf{NT-NEG} (Non-Targeted Negative): These comments contain hateful and offensive terms without any identifiable target.
\end{itemize}

\subsection{Annotation Process}
\label{subsec:Annotation_Process}

The annotation process for this study consists of three phases, involving five different annotators. In the first phase, the data were divided into five parts, and each part was annotated by two annotators. The Cohen's kappa score was then calculated between the two annotators. If the Cohen's kappa score was below 75\%, indicating poor agreement, the process moved to the second phase.

In the second phase, a consensus meeting was conducted to discuss the reasons behind the poor results. Based on the discussions, the annotation guidelines were modified as necessary. The annotators then re-labeled the data on which they had disagreed in the previous phase.

Once the inter-annotator agreement reached a minimum threshold of 75\%, the process moved to the third phase. In this phase, any remaining points of disagreement were resolved by a third annotator. The final label for each data point was determined based on the majority vote among the annotators.

\begin{figure}[htbp]
        \centering
        \includegraphics[width=\linewidth]{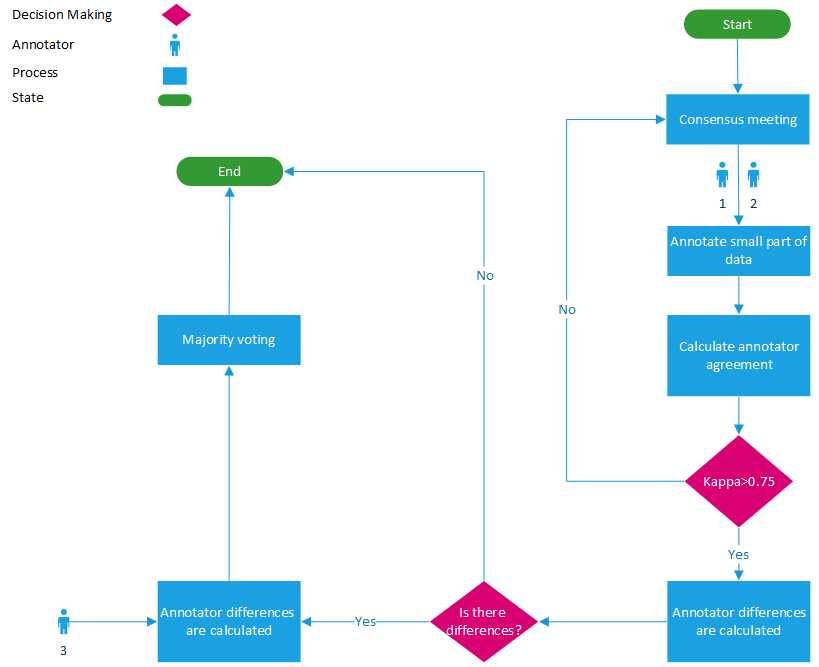}
        \caption{The annotation process}
        \label{fig:annotation_process}
\end{figure}

Figure \ref{fig:annotation_process} provides an overview of the annotation process.

\subsection{Annotation Reliability}
\label{subsec:Annotation_Reliability}
The annotation reliability in this study was assessed using Cohen's kappa coefficient. Cohen's kappa measures the agreement between annotators beyond what would be expected by chance.

Cohen's kappa is calculated as follows:
\begin{equation}
\kappa = \frac{p_o - p_e}{1 - p_e}
\end{equation}
where $p_o$ is the percentage of observed agreement between annotators and $p_e$ is the percentage of agreement expected by chance.

Two cases of labels were considered for calculating Cohen’s kappa. The first case involved the 3-class labeling, where each class of data was assigned a unique label. For the NEG+ case, which was used for binary classification, the NT-NEG and T-NEG labels were combined.

It is important to note that we combined two separate OPSD datasets, namely OPSD (Twitter) and OPSD (Instagram), and subsequently annotated the combined OPSD dataset.

Table \ref{tab:kappa-total} provides the Kappa scores for the OPSD dataset. The Phase column represents the first and the second phase of the labeling process. The Part column refers to five parts of the dataset, where each was annotated by two different annotators. The Annotator column indicates the five different annotators involved in the labeling process. The next three columns denote the number of comments labelled for each class. The Total and Common columns show the number of comments in each part, and the number of comments for which the annotators agreed on the label, respectively. Additionally, the Difference column indicates the number of differences in labeling between the two annotators. Lastly, the last column shows the Kappa score, considering the three classes, for each part of the dataset.

\begin{table*}[htbp]
\centering
\begin{tabular}{|c|c|c|c|c|c|c|c|c|c|}
\hline
\multicolumn{1}{|l|}{Phase} & Part               & Annotator  & POS  & NT-NEG & T-NEG & Total                 & Common                & Difference           & $\kappa_{3-class}$     \\ \hline
\multirow{10}{*}{I}        & \multirow{2}{*}{A} & Annotator1 & 3262 & 126   & 923  & \multirow{2}{*}{4311} & \multirow{2}{*}{3761} & \multirow{2}{*}{550} & \multirow{2}{*}{0.681} \\ 
                          &                    & Annotator2 & 3151 & 239   & 921  &                       &                       &                      &                        \\ 
                          & \multirow{2}{*}{B} & Annotator2 & 3113 & 300   & 904  & \multirow{2}{*}{4317} & \multirow{2}{*}{3672} & \multirow{2}{*}{645} & \multirow{2}{*}{0.626} \\ 
                          &                    & Annotator3 & 3318 & 51    & 948  &                       &                       &                      &                        \\ 
                          & \multirow{2}{*}{C} & Annotator3 & 3138 & 141   & 960  & \multirow{2}{*}{4239} & \multirow{2}{*}{3696} & \multirow{2}{*}{543} & \multirow{2}{*}{0.698} \\
                          &                    & Annotator4 & 2900 & 40    & 1299 &                       &                       &                      &                        \\  
                          & \multirow{2}{*}{D} & Annotator4 & 2809 & 113   & 1075 & \multirow{2}{*}{3997} & \multirow{2}{*}{3611} & \multirow{2}{*}{386} & \multirow{2}{*}{0.769} \\ 
                          &                    & Annotator5 & 2973 & 158   & 866  &                       &                       &                      &                        \\ 
                          & \multirow{2}{*}{E} & Annotator5 & 3191 & 61    & 1049 & \multirow{2}{*}{4301} & \multirow{2}{*}{3924} & \multirow{2}{*}{377} & \multirow{2}{*}{0.781} \\ 
                          &                    & Annotator1 & 3122 & 118   & 1061 &                       &                       &                      &                        \\ \hline
\multirow{10}{*}{II}       & \multirow{2}{*}{A} & Annotator1 & 3230 & 129   & 952  & \multirow{2}{*}{4311} & \multirow{2}{*}{4095} & \multirow{2}{*}{216} & \multirow{2}{*}{0.874} \\ 
                          &                    & Annotator2 & 3178 & 189   & 944  &                       &                       &                      &                        \\ 
                          & \multirow{2}{*}{B} & Annotator2 & 3183 & 194   & 940  & \multirow{2}{*}{4317} & \multirow{2}{*}{4044} & \multirow{2}{*}{273} & \multirow{2}{*}{0.839} \\ 
                          &                    & Annotator3 & 3273 & 83    & 961  &                       &                       &                      &                        \\  
                          & \multirow{2}{*}{C} & Annotator3 & 3075 & 147   & 1017 & \multirow{2}{*}{4239} & \multirow{2}{*}{4121} & \multirow{2}{*}{118} & \multirow{2}{*}{0.933} \\
                          &                    & Annotator4 & 3069 & 150   & 1020 &                       &                       &                      &                        \\  
                          & \multirow{2}{*}{D} & Annotator4 & 2878 & 147   & 972  & \multirow{2}{*}{3997} & \multirow{2}{*}{3828} & \multirow{2}{*}{169} & \multirow{2}{*}{0.899} \\ 
                          &                    & Annotator5 & 2889 & 159   & 949  &                       &                       &                      &                        \\  
                          & \multirow{2}{*}{E} & Annotator5 & 3201 & 106   & 994  & \multirow{2}{*}{4301} & \multirow{2}{*}{4166} & \multirow{2}{*}{135} & \multirow{2}{*}{0.922} \\ 
                          &                    & Annotator1 & 3133 & 124   & 1044 &                       &                       &                      &                        \\ \hline
\end{tabular}
\caption{Kappa score of the OPSD dataset for each phase and each part}
\label{tab:kappa-total}
\end{table*}

The average agreement scores per annotation phase are presented in Table \ref{tab:kappa-total-agg}. This table provides information on the number of instances for each label in each phase, the number of agreements and disagreements between annotators, and the average kappa score for the five parts of the dataset.

\begin{table*}[htbp]
\centering
\begin{tabular}{c|ccc|c}
\hline
Phase & Total & Common & Difference & $\kappa_{3-class}$ \\ \hline
I     & 21165 & 18664  & 2501       & 0.711              \\ \hline
II    & 21165 & 20254  & 911        & 0.8934             \\ \hline
\end{tabular}
\caption{The number of each label and average Kappa scores of OPSD dataset in two phases of labeling process.}
\label{tab:kappa-total-agg}
\end{table*}

Table \ref{tab:sample_data}, presents a selection of example samples from the dataset, showcasing the Persian text associated with each label.

\begin{table*}[!ht]
\centering
\begin{tabular}{| p{0.15\linewidth} | p{0.8\linewidth} | }
\hline
Label                   & Comment \\ \hline
POS  & For about a year now, the internet quality of Irancell has deteriorated significantly, to the point that I no longer use it. \\ \hline
POS & What is this previous debt on the Hamrah Aval (MCI) bill? For each billing cycle that you have an internet plan, the bill shows the outstanding balance from the previous month, which you pay along with the current month's charges. \\ \hline
POS & Hamrah Aval (MCI) has become quite burdensome with its expensive packages lately ... Come on, have a little mercy on people's situation. \\ \hline

NT-NEG & I see the gathering of foolish rulers here, don't they have your national code to ask you to enter it? \\ \hline 
NT-NEG & Is your internet speed also sluggish? \\ \hline
NT-NEG & He's expelled his essence, yet you're still recharging; you pour out more of him, the stench of his putrefaction spreads, and it contaminates everything around. \\ \hline

T-NEG  & The shameless ones have deleted the unlimited package. \\ \hline 
T-NEG & I need to go to the bathroom; the package doesn't work without electricity. My phone's battery is at 4\%, the modem is off, and Irancell hasn't left any balance. In the parking lot, the gate doesn't open without power, and my car is inside the parking lot, and I can't do anything. Curse your head and your nonsense. \\ \hline
T-NEG & Your driving is awful, people of Tabriz... I don't know about other cities, but I hope it's not as bad as this. If it is, even the ISIS government would be better for you. \\ \hline
\end{tabular}
\caption{Examples of the OPSD dataset for each label.}
\label{tab:sample_data}
\end{table*}

\subsection{Data Analysis}
\label{subsec:Annotation_Results}

\begin{figure*}[htbp]
    \centering
    \includegraphics[width=0.7 \linewidth]{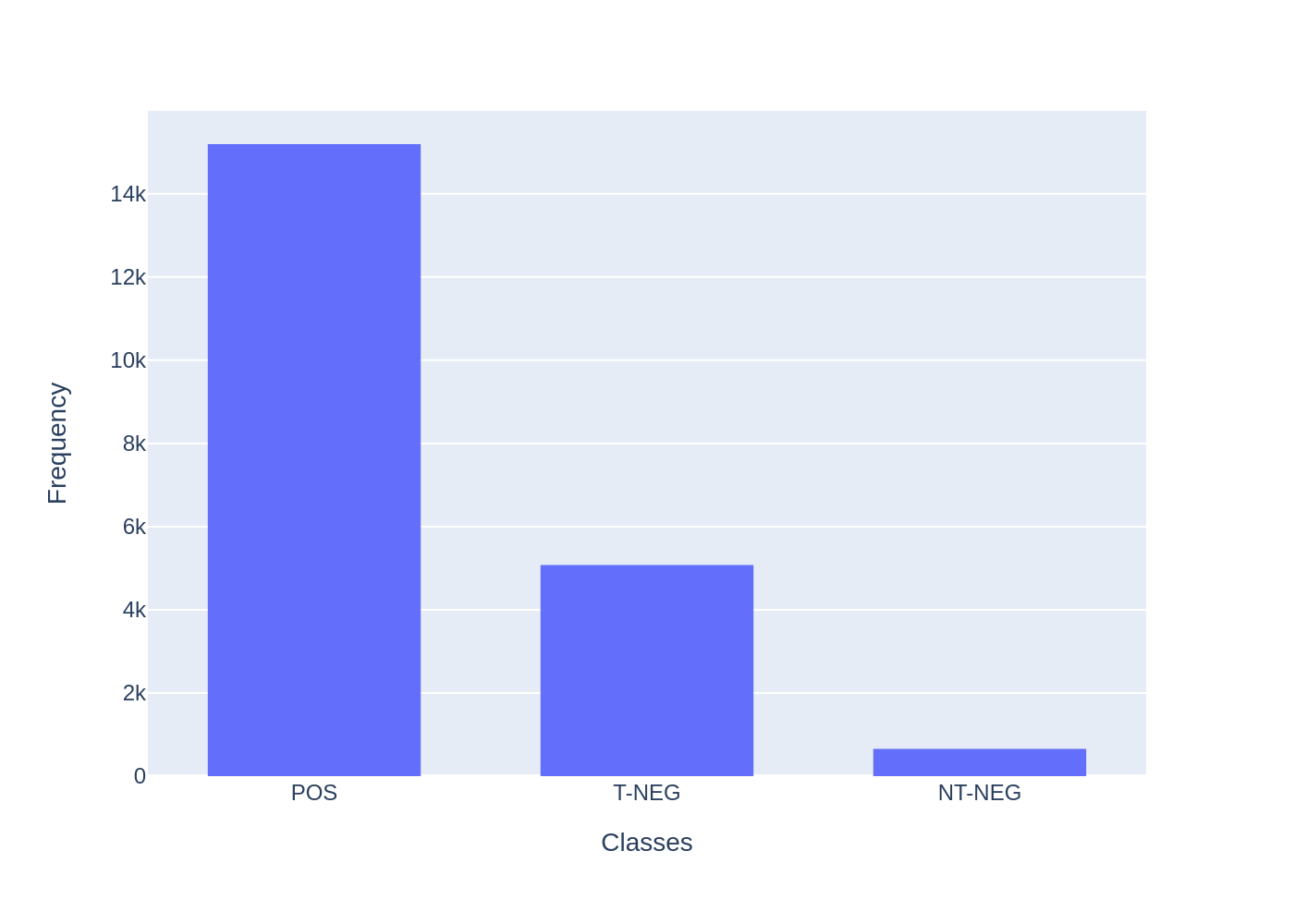} 
    \caption{The distribution of labels}
    \label{fig:words_dist}
\end{figure*}

As previously mentioned, a total of 21,165 comments were collected and categorized into Positive, Targeted Negative, and Non-Targeted Negative. The distribution of these tags among the comments is illustrated in Figure \ref{fig:words_dist} and the specific counts for each class are provided in Table \ref{tab:labels_dist}

\begin{table*}[htbp]
\centering
\begin{tabular}{lcc}
\hline
Class Name & Class Tag & Class Count \\ \hline
POS        & 1         & 15436         \\
NT-NEG     & 0         & 654         \\
T-NEG      & -1        & 5075         \\ \hline
\end{tabular}
\caption{Number of Each Class in the OPSD dataset with Class Tags.}
\label{tab:labels_dist}
\end{table*}

\begin{figure*}[htbp]
    \centering
    \includegraphics[width=0.8 \linewidth]{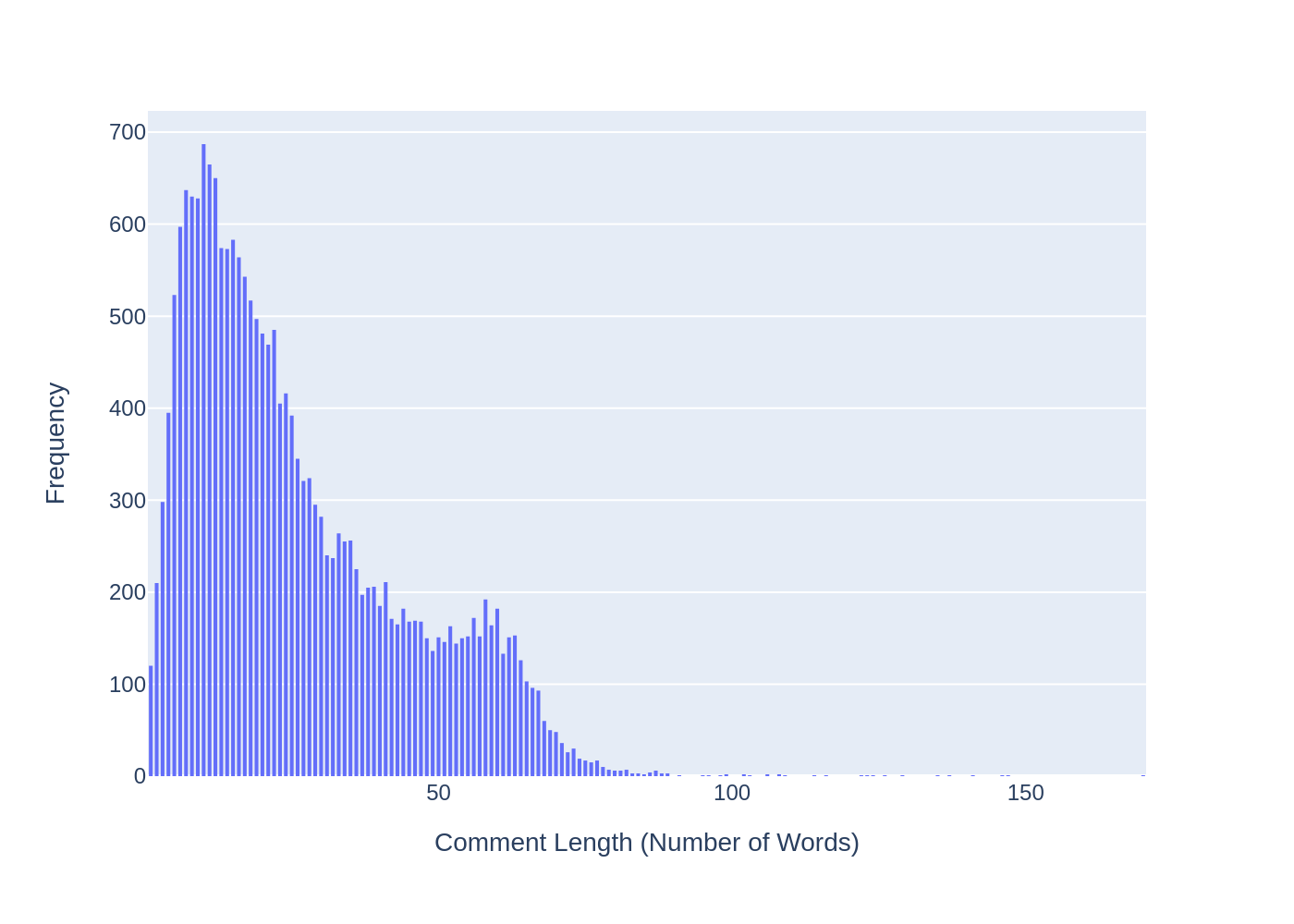} 
    \caption{Word-level distribution of comments length for the OPSD dataset.}
    \label{fig:comment_length_dist}
\end{figure*}

After a pre-processing step, which involved normalization and removal of numbers, URLs, email addresses, and special characters, the distribution of comments length (number of tokens) excluding emojis is presented in Figure \ref{fig:comment_length_dist}. The distribution of comment lengths reveals a prominent skew towards shorter texts, which coincides with the prevailing observation concerning comments in social media platforms.

\begin{figure*}[htbp]
    \centering
    \includegraphics[width=0.8 \linewidth]{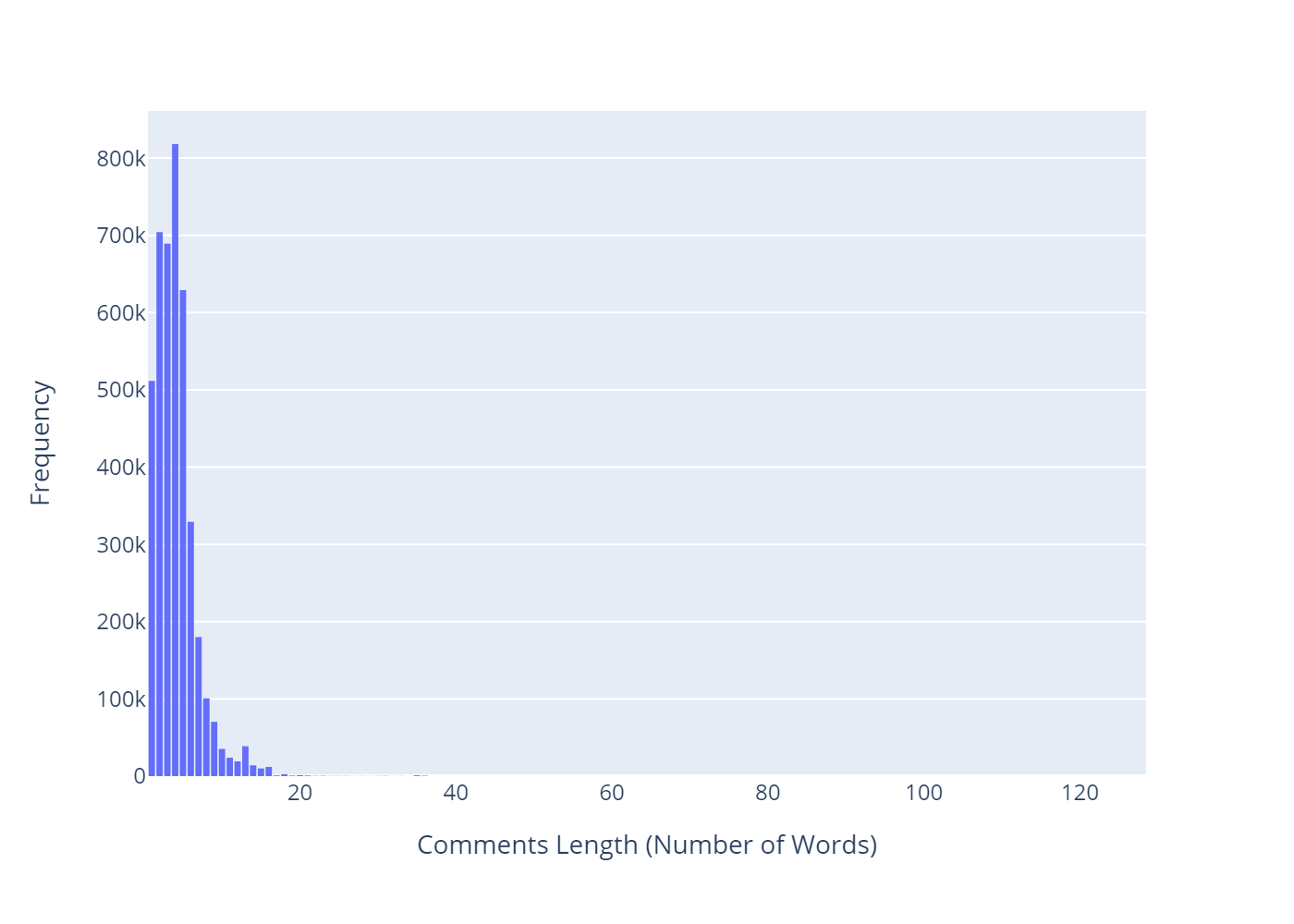} 
    \caption{Word-level distribution of comments length for the OPSD dataset (unlabeled).}
    \label{fig:comment_length_dist_unlabeled}
\end{figure*}

Moreover, the unprocessed OPSD (unlabeled) data also aligns with the same trend, as illustrated in Figure \ref{fig:comment_length_dist_unlabeled}.

\section{Experiments}
\label{sec:experiments}

We trained a classifier and compared multiple models using both supervised and unsupervised approaches. In supervised methods, we utilize machine learning and Transformer-based models, while in unsupervised methods, we employ a masked language model (MLM).

The dataset used in our classification experiments contains 21,165 entries and is split into a training set (80\%), development set (10\%), and test set (10\%). Hyperparameters are tuned on the development set.

We employed several machine learning methods, which served as baseline text categorization models. The feature vectors were constructed using the TF-IDF and Count Vectorizer methods. The results of applying these methods on the test data are presented in Table \ref{tab:restuls-tfidf}. 

The logistic regression \cite{scikit-learn} model was utilized with a regularization parameter (C) set to $1e-9$. K-Nearest Neighbors (KNN) algorithm \cite{scikit-learn} used 3 neighbors for classification, and the distance metric chosen was the Euclidean distance. We utilized the SVM algorithm \cite{scikit-learn} with a Radial Basis Function (RBF) kernel. The regularization parameter (C) was set to 0.025, and the gamma value was determined using the 'scale' method. We employed a single decision tree \cite{scikit-learn} for classification. The maximum depth of the tree was set to None, allowing nodes to expand until all leaves were pure or contained a minimum number of samples required for splitting, and the minimum samples required to split a node was set to 2. For the Random Forest Classifier \cite{scikit-learn}, we employed 200 decision trees as estimators. The maximum depth of the trees was not restricted (None), and the minimum number of samples required to split a node was set to 2. The Gaussian Naive Bayes algorithm \cite{scikit-learn} assumes that features follow a Gaussian distribution. It is particularly useful for continuous features. As it doesn't have any hyperparameters to tune, we used it with its default settings.

\begin{table*}[htbp]
\centering
\begin{tabular}{cccccc}
\hline
                                  &                        & \multicolumn{2}{c}{NEG+} & \multicolumn{2}{c}{3-Class} \\ \hline
Feature Extraction Method          & Model Name             & Acc         & F1         & Acc          & F1           \\ \hline
\multirow{6}{*}{Count Vectorizer}  & LogisticRegression     & 0.719       & 0.420      & 0.719        & 0.280        \\
                                  & KNeighborsClassifier   & 0.830       & 0.741      & 0.829        & 0.509        \\
                                  & SVC                    & 0.750       & 0.528      & 0.751        & 0.361        \\
                                  & DecisionTreeClassifier & 0.866       & 0.831      & 0.860        & 0.682        \\
                                  & RandomForestClassifier & 0.885       & 0.849      & 0.879        & 0.617        \\
                                  & GaussianNB             & 0.513       & 0.511      & 0.523        & 0.399        \\ \hline
\multirow{6}{*}{TF-IDF Vectorizer} & LogisticRegression     & 0.719       & 0.418      & 0.719        & 0.279        \\
                                  & KNeighborsClassifier   & 0.748       & 0.523      & 0.748        & 0.356        \\
                                  & SVC                    & 0.726       & 0.443      & 0.726        & 0.297        \\
                                  & DecisionTreeClassifier & 0.867       & 0.834      & 0.847        & 0.643        \\
                                  & RandomForestClassifier & 0.879       & 0.839      & 0.875        & 0.614        \\
                                  & GaussianNB             & 0.504       & 0.499      & 0.509        & 0.385        \\ \hline
\end{tabular}
\caption{Results of applying some machine learning methods on the test set}
\label{tab:restuls-tfidf}
\end{table*}

In our experimental analysis, we employed the following transformer-based models, including multilingual and monolingual models, to comprehensively explore their respective performances and capabilities.

\paragraph{ParsBERT}
ParsBERT \cite{ParsBERT} is a monolingual language model based on the BERT \cite{devlin-etal-2019-bert} architecture. It is pre-trained on more than 3.9 million Persian documents, allowing it to capture contextual representations of Persian text.

\paragraph{ALBERT}

ALBERT \cite{lan2020albert} is a "Lite" version of BERT designed to improve memory efficiency. It employs parameter-reduction techniques such as dividing the embedding matrix into smaller matrices and using shared parameters across different BERT hidden layers. ALBERT-fa \cite{ALBERTPersian} is a monolingual language model based on the ALBERT architecture, specifically trained on Persian data.

\paragraph{RoBERTa} RoBERTa \cite{zhuang-etal-2021-robustly} is a variant of the BERT model that employs a larger training dataset and longer training duration to capture more extensive language representations. It removes the next sentence prediction objective used in BERT and utilizes dynamic masking during pretraining. RoBERTa-fa \footnote{https://huggingface.co/HooshvareLab/roberta-fa-zwnj-base} is a variant of RoBERTa trained specifically on Persian data.

XLM-RoBERTa (XLM-R) \citep{conneau-etal-2020-unsupervised} is a variant of RoBERTa that incorporates cross-lingual language modeling capabilities. It is designed for multilingual tasks, enabling effective understanding and processing of text in multiple languages. XLM-RoBERTa leverages pre-training on large-scale multilingual corpora to capture cross-lingual knowledge.

For the fine-tuning of these models, we added a fully-connected layer to the pre-trained model. We used weighted cross-entropy loss as the objective function, a batch size of 32, a maximum sequence length of 128, a learning rate of $5e-6$, a dropout rate of 0.5, and the number of epochs 10. The AdamW \cite{DBLP:journals/corr/abs-1711-05101} optimizer with a linear learning rate scheduler (without warm-up steps) was used.

To further enhance the model's performance, we conducted an extensive hyperparameter tuning process. We experimented with various hyperparameter configurations, exploring batch size $\in \{8, 16, 32 \}$, maximum sequence length $\in \{64, 128 \}$, learning rate $\in \{2e-5, 1e-5, 9e-6, 5e-6, 1e-6\}$, and dropout rate in the range of $[0.3-0.8]$. Through this thorough grid search approach, we aimed to find the most suitable hyperparameters for our specific task.

After conducting the hyperparameter tuning, we found that the initial hyperparameter values provided the best performance on our evaluation metrics. Consequently, we decided to retain these hyperparameters for the final training of models.

To evaluate models on the dataset, we used a multilingual model and three monolingual models that are specifically pre-trained on Persian data. According to table \ref{tab:results_fine_tune}, XLM-RoBERTa has performed best accuracy and F1-score on the test data in both NEG+ and 3-class.

\begin{table*}[htbp]
\centering
\begin{tabular}{ccccc}
\hline
            & \multicolumn{2}{c}{NEG+}        & \multicolumn{2}{c}{3-Class}     \\
            & \multicolumn{1}{l}{Acc} & F1    & \multicolumn{1}{l}{Acc} & F1    \\ \hline
ALBERT-fa   & 0.910                   & 0.890 & 0.891                   & 0.763 \\
ParsBERT    & 0.904                   & 0.883 & 0.887                   & 0.726 \\
RoBERTa-fa  & 0.908                   & 0.885 & 0.902                   & 0.741 \\
XLM-RoBERTa & 0.917                   & 0.899 & 0.907                   & 0.769 \\ \hline
\end{tabular}
\caption{Results on the test set in fine-tuning phase for each pre-trained language model}
\label{tab:results_fine_tune}
\end{table*}

\paragraph{Masked Language Model}

In this section, we present an experiment conducted to investigate the effects of additional training of LLMs using OPSD (unlabeled) data. Specifically, the approach involved training the LLM with OPSD (unlabeled) through Masked Language Modeling (MLM) \footnote{The task of masking tokens in sequence of tokens using [MASK] token and leading the language model (LM) in a way that it fills this masked token with appropriate token is so called mask language modeling (MLM).}. Subsequently, the LLM was trained on Sentiment Analysis task (SA) using OPSD (labeled) data to observe potential improvements.

The ALBERT-fa, ParsBERT, RoBERTa-fa, and XLM-RoBERTa models underwent additional training using the following parameters: epoch=50, learning rate=2e-6, max sequence length=128, and batch sizes of 8, 16, 32, and 64 on MLM task. Subsequently, the proposed models were fine-tuned in the same manner as in previous experiments on the SA task, and the corresponding results are presented in Table \ref{tab:results_mlm}.

The comparison of results between Table \ref{tab:results_fine_tune} and Table \ref{tab:results_mlm} indicates an interesting fact. Further training the language models on the MLM task improved the performance of the proposed models, particularly in terms of the F1-score. Notably, the Roberta model fine-tuned on OPSD (3-class) showed a 4.06\% increase, as shown in table \ref{tab:results_mlm}.

\begin{table*}[htbp]
\centering
\begin{tabular}{ccccc}
\hline
            & \multicolumn{2}{c}{NEG+}         & \multicolumn{2}{c}{3-Class}      \\
            & \multicolumn{1}{l}{Acc} & F1     & \multicolumn{1}{l}{Acc} & F1     \\ \hline
ALBERT-fa   & 0.9259                  & 0.9087 & 0.9077                  & 0.7867 \\
ParsBERT    & 0.9245                  & 0.9065 & 0.9125                  & 0.7648 \\
RoBERTa-fa  & 0.9211                  & 0.9024 & 0.9106                  & 0.7812 \\
XLM-RoBERTa & 0.9278                  & 0.9111 & 0.9130                  & 0.7756 \\ \hline
\end{tabular}
\caption{Results on the test set in masked language model (MLM) phase for each pre-trained language model}
\label{tab:results_mlm}
\end{table*}

\subsection{Error Analysis of the Best Model (XLM-RoBERTa)}

\begin{figure*}[htbp]
    \centering
    \includegraphics[width=0.4 \linewidth]{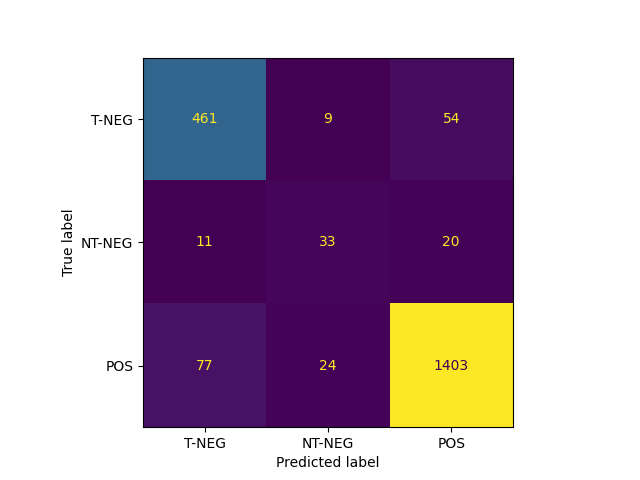} 
    \includegraphics[width=0.4 \linewidth]{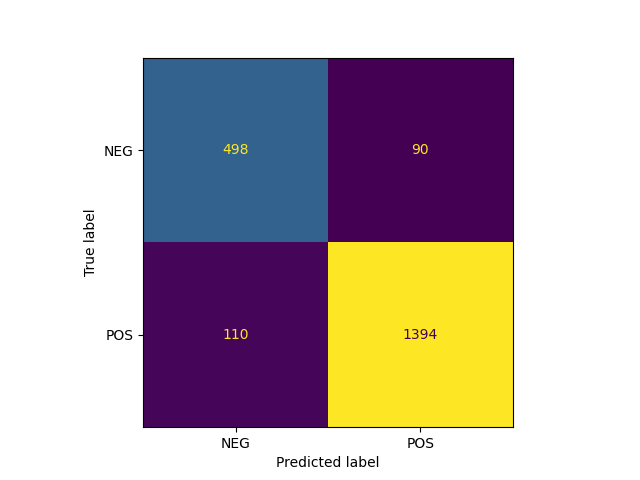} 
    \caption{Confusion matrices of the XLM-RoBERTa model. (left) 3-class case and (right) NEG+ case.}
    \label{fig:confusion-matrix}
\end{figure*}

To gain deeper insights into the performance of the best models in our experiment, we conduct an error analysis using their predictions. This analysis allows us to examine cases where the model might have made mistakes and provides valuable information for understanding its strengths and weaknesses. XLM-RoBERTa was chosen for this analysis due to its outstanding performance on the validation and test sets during our experimental evaluations. Figure \ref{fig:confusion-matrix} shows the confusion matrices of the XLM-RoBERTa model on the test set for both NEG+ and 3-class cases.

Error analysis for the 3-class case shows that misclassification on the test data can be categorized in two groups. The first group is the data samples that have POS labels, but the model predicts T-NEG or NT-NEG labels (FN) based on existence of some specific words. The observed issue arises from the discrepancy in the frequency of certain words within comments labeled as T-NEG and NT-NEG, as opposed to the significantly lower occurrence of these words in positive comments. This imbalance in word distribution could potentially lead to difficulties in effectively classifying comments with T-NEG and NT-NEG labels, as the model may not have encountered sufficient positive instances to learn from. Consequently, this disparity in the occurrence of specific words between different classes might contribute to the model's challenges in accurately distinguishing between the negative and positive comments.

The second group is the data samples that are not correctly labeled, even with high agreement between two annotators. This means that it seemed the model had predicted the labels correctly, but the data samples had been mislabeled. If we remove these data points from the test dataset, the results in Table \ref{tab:error analysis} are obtained, showing an approximately 3\% increase in accuracy for the XLM-RoBERTa model.

\begin{table}[hbt!]
\begin{tabular}{c|ccc}
\hline
              & \# Error & Accuracy & F1    \\ \hline
Before Removal & 195             & 0.907    & 0.769 \\ \hline
After Removal  & 139             & 0.936    & 0.842 \\ \hline
\end{tabular}
\caption{The performance of XLM-RoBERTa before and after removing mislabeled data. Number of errors is the total number of FP and FN.}
\label{tab:error analysis}
\end{table}

\section{Conclusion}
\label{sec:conclusion}

In this paper, we introduced OPSD, a dataset consisting of Twitter and Instagram comments written in Persian. The dataset is specifically designed to facilitate the detection of offensive language in text.

We applied a three-phase process for annotating the comments in the dataset manually. In the first phase, the dataset is divided into five parts, and each part is annotated by two annotators. If the agreement falls below 75\%, the process moves to the second phase. In the second phase, a meeting is conducted to discuss the low Kappa scores, leading to updates in annotation guidelines. The annotators then re-label the previously disagreed data. Finally, in the third phase, any remaining disagreements are resolved by a third annotator, determining the final label based on the majority vote among the annotators.

Our proposed dataset, OPSD, includes an unlabeled subdivision known as OPSD (Unlabeled). This particular subset was utilized for an additional training stage on the Masked Language Modeling task. The results demonstrated that incorporating this training step with OPSD (unlabeled) data prior to conducting the sentiment analysis task enhanced the performance of baseline text classification models.

We applied baseline text classification models, by using machine learning models and also state-of-the-work monolingual and multilingual BERT-based models, to show the effectiveness of the dataset.

In the future, there is a plan to increase the size of the OPSD dataset. This would require considering semi-supervised or unsupervised techniques to determine offensive scores and classify textual entities, as labeling all items would be time-consuming. Additionally, improving preprocessing modules, such as converting emojis to their Persian textual descriptors, can enhance performance. Considering language structures and exploring other approaches, such as deep learning methods, for determining probabilities (offensive scores) can further improve the proposed methodology.

\backmatter

\bmhead{Acknowledgments}

This publication was supported by grant No. RD-53-0105-0012 from the R\&D Center of Mobile Telecommunication Company of Iran (MCI) for advancing information and communications technologies.

\bigskip

\bibliography{sn-bibliography}


\begin{thebibliography}{15}
\ifx \bisbn   \undefined \def \bisbn  #1{ISBN #1}\fi
\ifx \binits  \undefined \def \binits#1{#1}\fi
\ifx \bauthor  \undefined \def \bauthor#1{#1}\fi
\ifx \batitle  \undefined \def \batitle#1{#1}\fi
\ifx \bjtitle  \undefined \def \bjtitle#1{#1}\fi
\ifx \bvolume  \undefined \def \bvolume#1{\textbf{#1}}\fi
\ifx \byear  \undefined \def \byear#1{#1}\fi
\ifx \bissue  \undefined \def \bissue#1{#1}\fi
\ifx \bfpage  \undefined \def \bfpage#1{#1}\fi
\ifx \blpage  \undefined \def \blpage #1{#1}\fi
\ifx \burl  \undefined \def \burl#1{\textsf{#1}}\fi
\ifx \doiurl  \undefined \def \doiurl#1{\url{https://doi.org/#1}}\fi
\ifx \betal  \undefined \def \betal{\textit{et al.}}\fi
\ifx \binstitute  \undefined \def \binstitute#1{#1}\fi
\ifx \binstitutionaled  \undefined \def \binstitutionaled#1{#1}\fi
\ifx \bctitle  \undefined \def \bctitle#1{#1}\fi
\ifx \beditor  \undefined \def \beditor#1{#1}\fi
\ifx \bpublisher  \undefined \def \bpublisher#1{#1}\fi
\ifx \bbtitle  \undefined \def \bbtitle#1{#1}\fi
\ifx \bedition  \undefined \def \bedition#1{#1}\fi
\ifx \bseriesno  \undefined \def \bseriesno#1{#1}\fi
\ifx \blocation  \undefined \def \blocation#1{#1}\fi
\ifx \bsertitle  \undefined \def \bsertitle#1{#1}\fi
\ifx \bsnm \undefined \def \bsnm#1{#1}\fi
\ifx \bsuffix \undefined \def \bsuffix#1{#1}\fi
\ifx \bparticle \undefined \def \bparticle#1{#1}\fi
\ifx \barticle \undefined \def \barticle#1{#1}\fi
\bibcommenthead
\ifx \bconfdate \undefined \def \bconfdate #1{#1}\fi
\ifx \botherref \undefined \def \botherref #1{#1}\fi
\ifx \url \undefined \def \url#1{\textsf{#1}}\fi
\ifx \bchapter \undefined \def \bchapter#1{#1}\fi
\ifx \bbook \undefined \def \bbook#1{#1}\fi
\ifx \bcomment \undefined \def \bcomment#1{#1}\fi
\ifx \oauthor \undefined \def \oauthor#1{#1}\fi
\ifx \citeauthoryear \undefined \def \citeauthoryear#1{#1}\fi
\ifx \endbibitem  \undefined \def \endbibitem {}\fi
\ifx \bconflocation  \undefined \def \bconflocation#1{#1}\fi
\ifx \arxivurl  \undefined \def \arxivurl#1{\textsf{#1}}\fi
\csname PreBibitemsHook\endcsname

\bibitem[\protect\citeauthoryear{Zampieri et~al.}{2019}]{zampierietal2019}
\begin{bchapter}
\bauthor{\bsnm{Zampieri}, \binits{M.}},
\bauthor{\bsnm{Malmasi}, \binits{S.}},
\bauthor{\bsnm{Nakov}, \binits{P.}},
\bauthor{\bsnm{Rosenthal}, \binits{S.}},
\bauthor{\bsnm{Farra}, \binits{N.}},
\bauthor{\bsnm{Kumar}, \binits{R.}}:
\bctitle{{Predicting the Type and Target of Offensive Posts in Social Media}}.
In: \bbtitle{Proceedings of NAACL}
(\byear{2019})
\end{bchapter}
\endbibitem

\bibitem[\protect\citeauthoryear{Rosenthal
  et~al.}{2021}]{rosenthal-etal-2021-solid}
\begin{bchapter}
\bauthor{\bsnm{Rosenthal}, \binits{S.}},
\bauthor{\bsnm{Atanasova}, \binits{P.}},
\bauthor{\bsnm{Karadzhov}, \binits{G.}},
\bauthor{\bsnm{Zampieri}, \binits{M.}},
\bauthor{\bsnm{Nakov}, \binits{P.}}:
\bctitle{{SOLID}: A large-scale semi-supervised dataset for offensive language
  identification}.
In: \bbtitle{Findings of the Association for Computational Linguistics:
  ACL-IJCNLP 2021},
pp. \bfpage{915}--\blpage{928}.
\bpublisher{Association for Computational Linguistics},
\blocation{Online}
(\byear{2021}).
\doiurl{10.18653/v1/2021.findings-acl.80} .
\burl{https://aclanthology.org/2021.findings-acl.80}
\end{bchapter}
\endbibitem

\bibitem[\protect\citeauthoryear{Davidson
  et~al.}{2017}]{Davidson_Warmsley_Macy_Weber_2017}
\begin{barticle}
\bauthor{\bsnm{Davidson}, \binits{T.}},
\bauthor{\bsnm{Warmsley}, \binits{D.}},
\bauthor{\bsnm{Macy}, \binits{M.}},
\bauthor{\bsnm{Weber}, \binits{I.}}:
\batitle{Automated hate speech detection and the problem of offensive
  language}.
\bjtitle{Proceedings of the International AAAI Conference on Web and Social
  Media}
\bvolume{11}(\bissue{1}),
\bfpage{512}--\blpage{515}
(\byear{2017})
\doiurl{10.1609/icwsm.v11i1.14955}
\end{barticle}
\endbibitem

\bibitem[\protect\citeauthoryear{Alavi et~al.}{2021}]{alavi2021offensive}
\begin{botherref}
\oauthor{\bsnm{Alavi}, \binits{P.}},
\oauthor{\bsnm{Nikvand}, \binits{P.}},
\oauthor{\bsnm{Shamsfard}, \binits{M.}}:
Offensive language detection with bert-based models, by customizing attention
  probabilities.
arXiv preprint arXiv:2110.05133
(2021)
\end{botherref}
\endbibitem

\bibitem[\protect\citeauthoryear{Mozafari}{2021}]{mozafari:tel-03276023}
\begin{botherref}
\oauthor{\bsnm{Mozafari}, \binits{M.}}:
{Hate speech and offensive language detection using transfer learning
  approaches}.
Theses,
{Institut Polytechnique de Paris}
(May 2021).
\url{https://theses.hal.science/tel-03276023}
\end{botherref}
\endbibitem

\bibitem[\protect\citeauthoryear{Khodabakhsh et~al.}{2023}]{perbold}
\begin{barticle}
\bauthor{\bsnm{Khodabakhsh}, \binits{M.}},
\bauthor{\bsnm{Jafarinejad}, \binits{F.}},
\bauthor{\bsnm{Rahimi}, \binits{M.}},
\bauthor{\bsnm{Ghayoomi}, \binits{M.}}:
\batitle{Perbold: A big dataset of persian offensive language on instagram
  comments}.
\bjtitle{TABRIZ JOURNAL OF ELECTRICAL ENGINEERING}
\bvolume{53}(\bissue{2}),
\bfpage{149}--\blpage{158}
(\byear{2023})
\doiurl{10.22034/tjee.2023.15794}
{\href{https://arxiv.org/abs/https://tjee.tabrizu.ac.ir/article\_15794\_1a75a5b450027a6f41f6dfb9e8cf3e51.pdf}{{https://tjee.tabrizu.ac.ir/article\_15794\_1a75a5b450027a6f41f6dfb9e8cf3e51.pdf}}}
\end{barticle}
\endbibitem

\bibitem[\protect\citeauthoryear{Ataei et~al.}{2022}]{9936700}
\begin{botherref}
\oauthor{\bsnm{Ataei}, \binits{T.S.}},
\oauthor{\bsnm{Darvishi}, \binits{K.}},
\oauthor{\bsnm{Javdan}, \binits{S.}},
\oauthor{\bsnm{Pourdabiri}, \binits{A.}},
\oauthor{\bsnm{Minaei-Bidgoli}, \binits{B.}},
\oauthor{\bsnm{Pilehvar}, \binits{M.T.}}:
Pars-off: A benchmark for offensive language detection on farsi social media.
IEEE Transactions on Affective Computing,
1--9
(2022)
\doiurl{10.1109/TAFFC.2022.3219229}
\end{botherref}
\endbibitem

\bibitem[\protect\citeauthoryear{Pedregosa et~al.}{2011}]{scikit-learn}
\begin{barticle}
\bauthor{\bsnm{Pedregosa}, \binits{F.}},
\bauthor{\bsnm{Varoquaux}, \binits{G.}},
\bauthor{\bsnm{Gramfort}, \binits{A.}},
\bauthor{\bsnm{Michel}, \binits{V.}},
\bauthor{\bsnm{Thirion}, \binits{B.}},
\bauthor{\bsnm{Grisel}, \binits{O.}},
\bauthor{\bsnm{Blondel}, \binits{M.}},
\bauthor{\bsnm{Prettenhofer}, \binits{P.}},
\bauthor{\bsnm{Weiss}, \binits{R.}},
\bauthor{\bsnm{Dubourg}, \binits{V.}},
\bauthor{\bsnm{Vanderplas}, \binits{J.}},
\bauthor{\bsnm{Passos}, \binits{A.}},
\bauthor{\bsnm{Cournapeau}, \binits{D.}},
\bauthor{\bsnm{Brucher}, \binits{M.}},
\bauthor{\bsnm{Perrot}, \binits{M.}},
\bauthor{\bsnm{Duchesnay}, \binits{E.}}:
\batitle{Scikit-learn: Machine learning in {P}ython}.
\bjtitle{Journal of Machine Learning Research}
\bvolume{12},
\bfpage{2825}--\blpage{2830}
(\byear{2011})
\end{barticle}
\endbibitem

\bibitem[\protect\citeauthoryear{Farahani et~al.}{2021}]{ParsBERT}
\begin{barticle}
\bauthor{\bsnm{Farahani}, \binits{M.}},
\bauthor{\bsnm{Gharachorloo}, \binits{M.}},
\bauthor{\bsnm{Farahani}, \binits{M.}},
\bauthor{\bsnm{Manthouri}, \binits{M.}}:
\batitle{Parsbert: Transformer-based model for persian language understanding}.
\bjtitle{Neural Processing Letters}
\bvolume{53}(\bissue{6}),
\bfpage{3831}--\blpage{3847}
(\byear{2021})
\doiurl{10.1007/s11063-021-10528-4}
\end{barticle}
\endbibitem

\bibitem[\protect\citeauthoryear{Devlin et~al.}{2019}]{devlin-etal-2019-bert}
\begin{bchapter}
\bauthor{\bsnm{Devlin}, \binits{J.}},
\bauthor{\bsnm{Chang}, \binits{M.-W.}},
\bauthor{\bsnm{Lee}, \binits{K.}},
\bauthor{\bsnm{Toutanova}, \binits{K.}}:
\bctitle{{BERT}: Pre-training of deep bidirectional transformers for language
  understanding}.
In: \bbtitle{Proceedings of the 2019 Conference of the North {A}merican Chapter
  of the Association for Computational Linguistics: Human Language
  Technologies, Volume 1 (Long and Short Papers)},
pp. \bfpage{4171}--\blpage{4186}.
\bpublisher{Association for Computational Linguistics},
\blocation{Minneapolis, Minnesota}
(\byear{2019}).
\doiurl{10.18653/v1/N19-1423} .
\burl{https://aclanthology.org/N19-1423}
\end{bchapter}
\endbibitem

\bibitem[\protect\citeauthoryear{Lan et~al.}{2020}]{lan2020albert}
\begin{botherref}
\oauthor{\bsnm{Lan}, \binits{Z.}},
\oauthor{\bsnm{Chen}, \binits{M.}},
\oauthor{\bsnm{Goodman}, \binits{S.}},
\oauthor{\bsnm{Gimpel}, \binits{K.}},
\oauthor{\bsnm{Sharma}, \binits{P.}},
\oauthor{\bsnm{Soricut}, \binits{R.}}:
ALBERT: A Lite BERT for Self-supervised Learning of Language Representations
(2020)
\end{botherref}
\endbibitem

\bibitem[\protect\citeauthoryear{Team}{2021}]{ALBERTPersian}
\begin{botherref}
\oauthor{\bsnm{Team}, \binits{H.}}:
ALBERT-Persian: A Lite BERT for Self-supervised Learning of Language
  Representations for the Persian Language.
GitHub
(2021)
\end{botherref}
\endbibitem

\bibitem[\protect\citeauthoryear{Zhuang
  et~al.}{2021}]{zhuang-etal-2021-robustly}
\begin{bchapter}
\bauthor{\bsnm{Zhuang}, \binits{L.}},
\bauthor{\bsnm{Wayne}, \binits{L.}},
\bauthor{\bsnm{Ya}, \binits{S.}},
\bauthor{\bsnm{Jun}, \binits{Z.}}:
\bctitle{A robustly optimized {BERT} pre-training approach with post-training}.
In: \bbtitle{Proceedings of the 20th Chinese National Conference on
  Computational Linguistics},
pp. \bfpage{1218}--\blpage{1227}.
\bpublisher{Chinese Information Processing Society of China},
\blocation{Huhhot, China}
(\byear{2021}).
\burl{https://aclanthology.org/2021.ccl-1.108}
\end{bchapter}
\endbibitem

\bibitem[\protect\citeauthoryear{Conneau
  et~al.}{2020}]{conneau-etal-2020-unsupervised}
\begin{bchapter}
\bauthor{\bsnm{Conneau}, \binits{A.}},
\bauthor{\bsnm{Khandelwal}, \binits{K.}},
\bauthor{\bsnm{Goyal}, \binits{N.}},
\bauthor{\bsnm{Chaudhary}, \binits{V.}},
\bauthor{\bsnm{Wenzek}, \binits{G.}},
\bauthor{\bsnm{Guzm{\'a}n}, \binits{F.}},
\bauthor{\bsnm{Grave}, \binits{E.}},
\bauthor{\bsnm{Ott}, \binits{M.}},
\bauthor{\bsnm{Zettlemoyer}, \binits{L.}},
\bauthor{\bsnm{Stoyanov}, \binits{V.}}:
\bctitle{Unsupervised cross-lingual representation learning at scale}.
In: \bbtitle{Proceedings of the 58th Annual Meeting of the Association for
  Computational Linguistics},
pp. \bfpage{8440}--\blpage{8451}.
\bpublisher{Association for Computational Linguistics},
\blocation{Online}
(\byear{2020}).
\doiurl{10.18653/v1/2020.acl-main.747} .
\burl{https://aclanthology.org/2020.acl-main.747}
\end{bchapter}
\endbibitem

\bibitem[\protect\citeauthoryear{Loshchilov and
  Hutter}{2017}]{DBLP:journals/corr/abs-1711-05101}
\begin{botherref}
\oauthor{\bsnm{Loshchilov}, \binits{I.}},
\oauthor{\bsnm{Hutter}, \binits{F.}}:
Fixing weight decay regularization in adam.
CoRR
\textbf{abs/1711.05101}
(2017)
{\href{https://arxiv.org/abs/1711.05101}{{1711.05101}}}
\end{botherref}
\endbibitem

\end{thebibliography}

\end{document}